\documentclass[letterpaper, conference, 10pt, final]{ieeeconf}
\IEEEoverridecommandlockouts
\usepackage{cite}
\usepackage{amsmath,amssymb,amsfonts}
\usepackage{algorithmic}
\usepackage{graphicx}
\usepackage{textcomp}
\usepackage{xcolor}
\usepackage{mathtools}
\usepackage{tikz}

\newcommand\copyrighttext{%
  \footnotesize \copyright~2024 IEEE. Personal use of this material is permitted.  Permission from IEEE must be obtained for all other uses, in any current or future media, including reprinting/republishing this material for advertising or promotional purposes, creating new collective works, for resale or redistribution to servers or lists, or reuse of any copyrighted component of this work in other works.}
\newcommand\copyrightnotice{%
\begin{tikzpicture}[remember picture,overlay]
\node[anchor=south,yshift=10pt] at (current page.south) {\parbox{\dimexpr\textwidth-\fboxsep-\fboxrule\relax}{\copyrighttext}};
\end{tikzpicture}%
}

\begin{document}

\title{\LARGE \bf
Inverse Kinematics for Neuro-Robotic Grasping with Humanoid Embodied Agents
}

\author{Jan-Gerrit~Habekost$^1$, Connor~Gäde$^1$, Philipp~Allgeuer$^1$, and Stefan~Wermter$^1$
\thanks{The authors gratefully acknowledge the support from the
DFG under project CML and from the BMWK under project VERIKAS.}
\thanks{$^1$ J.-G. Habekost, C. Gäde, P. Allgeuer, and S. Wermter are with the Knowledge Technology Group, Department of Informatics, University of Hamburg, Hamburg, Germany,
e-mail: \{jan-gerrit.habekost, stefan.wermter\}@uni-hamburg.de}}%

\maketitle

\begin{abstract}

This paper introduces a novel zero-shot motion planning method that allows users to quickly design smooth robot motions in Cartesian space. A B\'ezier curve-based Cartesian plan is transformed into a joint space trajectory by our neuro-inspired inverse kinematics (IK) method CycleIK, for which we enable platform independence by scaling it to arbitrary robot designs. The motion planner is evaluated on the physical hardware of the two humanoid robots NICO and NICOL in a human-in-the-loop grasping scenario. Our method is deployed with an embodied agent that is a large language model (LLM) at its core. We generalize the embodied agent, that was introduced for NICOL, to also embody NICO. The agent can execute a discrete set of physical actions and allows the user to verbally instruct various different robots. We contribute a grasping primitive to its action space that allows for precise manipulation of household objects. The updated CycleIK\footnote[2]{Available on GitHub: https://github.com/jangerritha/cycleik} method is compared to popular numerical IK solvers and state-of-the-art neural IK methods in simulation and is shown to be competitive with or outperform all evaluated methods when the algorithm runtime is very short. The grasping primitive is evaluated on both NICOL and NICO robots with a reported grasp success of 72\% to 82\% for each robot, respectively.

\end{abstract}

\copyrightnotice

\vspace{-5pt}
\section{Introduction}
Precise inverse kinematics is the key factor for successful grasping motions in many robotic manipulation applications. IK methods predict joint angles that align the robot end effector with a given position and orientation. They therefore collapse the action space of higher layered motion planners to the 6D pose of the end effector, while the alternative is to control the robot directly in its joint space which poorly scales for robots with varying degrees of freedom (DoF). The complexity of the IK task increases drastically for redundant kinematic chains with more than six DoF, since for non-redundant chains a specific pose maps exactly to one joint configuration, while for redundant manipulators infinite solutions generally exist, which makes the problem ambiguous. The number of neuro-inspired IK methods that are precise enough to be applied to physical hardware is increasing but there are only a few approaches that utilize them for neural motion planning\cite{MorganCppFlow:Planning, Park2022NODEIK:Planning}. Many motion planners exist that use neural networks to sample the robot joint space and deploy classical planning algorithms like MPNet\cite{Qureshi2021MotionPlanners}. Thus, only the runtime of the sampling decreases, while the latency for the motion planning itself stays high. Other approaches plan directly in Cartesian space and transform the plan to the joint space via inverse kinematics such as CppFlow\cite{MorganCppFlow:Planning}. Neural networks that learn to solve the IK task, which allows for parallel batched queries, can dramatically accelerate Cartesian planning methods but decrease the precision. Numerical IK approaches in difference have to solve each pose in an individual optimization run, resulting in a trade-off between precision and runtime, for which we propose one of the most runtime-efficient motion planning systems available. 

\begin{figure}[t!]
        \vspace{12pt}
      \centering
      \includegraphics[width=1.0\linewidth]{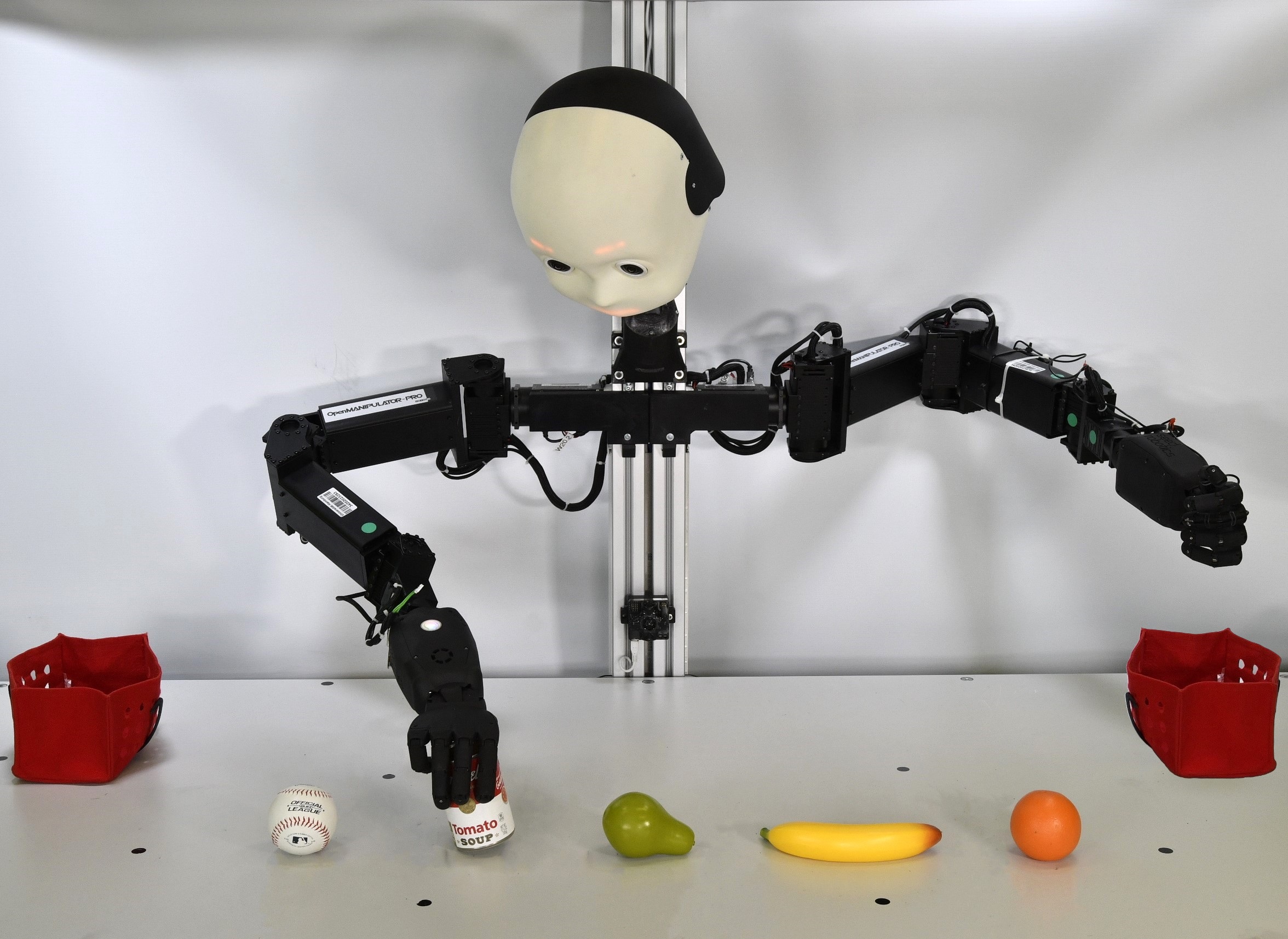}
      \caption{Humanoid embodied agent grasping scenario deployed to the physical NICOL robot.}
      \label{fig:nicol}
      \vspace{6pt}
\end{figure}

The neural inverse kinematics solver CycleIK\cite{10.1007/978-3-031-44207-0_38} is the central component of our proposed motion planning approach. CycleIK is scaled to arbitrary robot designs by a replacement of the loss functions for the positional and rotational error with two new metrics that are derived from the Smooth L1 loss\cite{Girshick2015FastR-CNN}. The new loss functions enable high precision within ten epochs of training, while our former publication suggested an ideal training time of $100$ to $300$ epochs using the old loss functions. The low-level motion planner interpolates a B\'ezier curve between the current pose and a target pose and then converts the Cartesian plan into a joint space trajectory in a single step via CycleIK. Our method is deployed with the high-level embodied agent that is introduced to the NICOL\cite{Kerzel2023NICOL:Manipulation} robot by Allgeuer et al.\cite{Allgeuer2024TBAAgent}. We generalize the embodied agent architecture over our humanoid robot platform to also embody the NICO\cite{Kerzel2017NICO-Neuro-inspiredInteraction} robot. 

The humanoid embodied agent introduced by Allgeuer et al.\cite{Allgeuer2024TBAAgent} is controlled with OpenAI GPT-3.5\cite{OpenAI2024OpenAIgpt-3.5-turbo-0301} and combines various neural architectures for different domains to create an interactive human-in-the-loop setup. The open-vocabulary object detector ViLD\cite{Gu2021Open-vocabularyDistillation} is utilized to detect the objects in front of the robots, and OpenAI Whisper\cite{Radford2022RobustSupervision} is used to recognize the verbal instructions of the user.

The updated CycleIK methodology is evaluated against two well-established IK methods available in Moveit\cite{Coleman2017ReducingStudy}, the genetic algorithm BioIK\cite{Starke2017AMotion} and the Jacobian-based method Trac-IK\cite{Beeson2015TRAC-IK:Kinematics}. The algorithms are compared on five different robot platforms: NICO, NICOL, Franka Emika Panda\cite{Haddadin2022TheEducation}, NASA Valkyrie\cite{Radford2015Valkyrie:Robot}, and the Fetch\cite{Wise2016FetchApplications} robot. In addition, the proposed IK method is compared against two state-of-the-art neuro-generative IK approaches by Limoyo et al.\cite{Limoyo2023EuclideanKinematics} and Ames et al.\cite{Ames2022IKFlow:Solutions}. Finally, the grasp success of our method is evaluated on the NICO and the NICOL robots, for $100$ grasp attempts on each robot.

\section{Related Work}

Invertible Neural Networks (INNs) were introduced by Ardizzone et al.\cite{Ardizzone2018AnalyzingNetworks} for ambiguous inverse problems and became a popular method for neuro-inspired inverse kinematics. However, the authors only evaluate their architecture on a three-DoF robot arm. IKFlow\cite{Ames2022IKFlow:Solutions} is based on the findings of \cite{Ardizzone2018AnalyzingNetworks} but generalizes it to various redundant manipulator designs. It is one of the most precise neural IK methods available and is capable of sampling the null space. Kim and Perez\cite{Kim2021LearningManipulator} propose a very similar setup that also utilizes normalizing flow-based INNs. In contrast, the FK model is not based on the robot URDF but is approximated by a separate neural network. The authors report very high errors on the precision. NodeIK\cite{Park2022NODEIK:Planning} is another conditioned normalizing flow IK method that is very similar to IKFlow, which encodes the latent space as neural ordinary differential equations. INNs show high performance for the IK task, but our results\cite{10.1007/978-3-031-44207-0_38} indicate that architectures that consist purely of fully connected layers can reach the same precision.

There are various approaches to neuro-generative inverse kinematics, but only a few reach a high precision. Limoyo et al.\cite{Limoyo2023EuclideanKinematics} introduce a precise graph neural network (GNN) approach that uses an encoder GNN to embed the robot state and the target pose into a latent space representation that is used by a decoder GNN to analytically compute the joint angles. In contrast, GNNs typically have a higher runtime than INNs and MLPs. 
Furthermore, Lembono et al.\cite{Lembono2020LearningNetwork} evaluate multiple variants of generative adversarial networks that are capable of sampling the poses null space but report quite large errors. IKNet\cite{Bensadoun2022NeuralKinematic} introduces a hierarchical neural network approach in which a hypernet parametrizes the fully connected IK network for each joint independently, conditioned to the input pose. Wagaa et al.\cite{Wagaa2023AnalyticalArm} test different variants of recurrent LSTM and GRU architectures to solve the IK task for non-redundant six-DoF robots. 

Neuro-inspired animation frameworks solve the IK problem but are not directly applicable to robotics, as they often do not require high precision. ProtoRes\cite{Oreshkin2021ProtoRes:Kinematics} is an animation framework that generates character motions from sparse user inputs. The user input is encoded into a latent embedding, which is decoded to joint angles by an IK network. SMPL-IK\cite{Voleti2022SMPL-IK:Workflows} extends ProtoRes which aligns the character pose with the pose estimation of a 2D input image of a human. The end effector poses from the human pose estimation are transferred to the character end effectors, then inverse kinematics are solved by a neural network. HybrIK\cite{Li2021HybrIK:Estimation} is an inverse kinematics solver for human pose estimation. The pose key points of an input image are extracted with a convolutional neural network and then a fully connected network, that takes the key points as input, predicts the joint angles. NIKI\cite{Li2023NIKI:Estimation} is an INN-based human pose estimation method. IK and FK are trained jointly, similar to IKFlow.

\section{Method}
The presented method fundamentally updates our neuro-inspired IK approach CycleIK\cite{10.1007/978-3-031-44207-0_38} to enable platform independence for our method. Most importantly, the loss functions for the positional and rotational errors are exchanged by new Smooth L1-based metrics. The CycleIK multi-layer perceptron (MLP) is a single-solution IK solver that deterministically predicts a single vector of joint angles for a specific pose. Secondary constraints, such as zero-joint goals or pose goals for other links in the kinematic chain, can optionally be applied during the training to influence the kinematic behavior of the model. CycleIK also provides a generative adversarial network that is capable of sampling the null space of a specific pose for redundant kinematic chains with higher than six DoF. The GAN can enable obstacle-avoiding motion planning, but as this study does not consider obstacle-avoidance, the MLP is sufficient since it is more applicable and will therefore be used to solve the inverse kinematics problem in our experiments.

\begin{figure*}[ht]
\begin{minipage}[b]{0.48\linewidth}
\centering
\includegraphics[width=\textwidth]{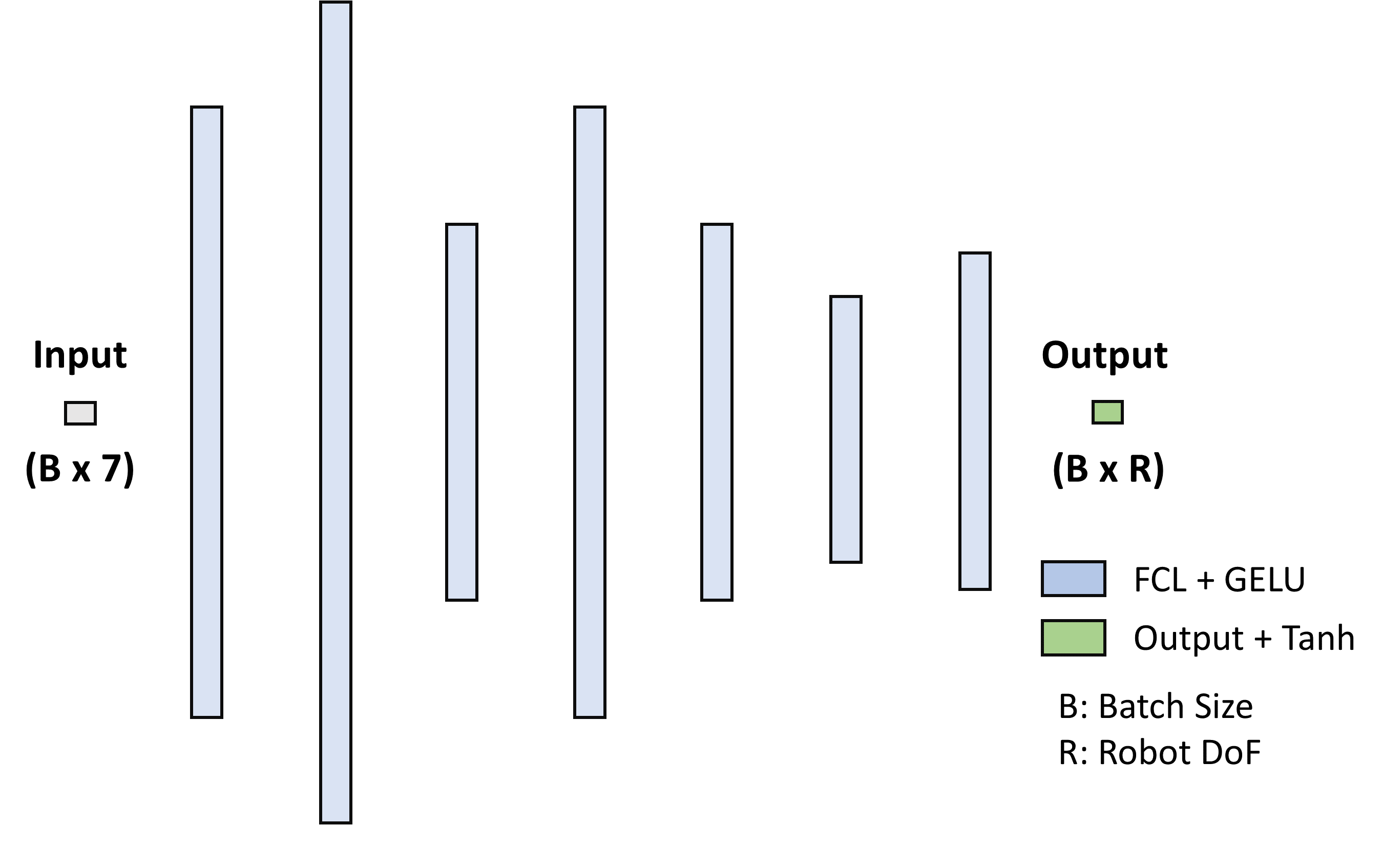}
\caption{The architecture of the CycleIK MLP. The hidden layers have the GELU\cite{Hendrycks2016GaussianGELUs} activation and the output has the tanh activation.}
\label{fig:cycleik_architecture}
\end{minipage}
\hspace{0.35cm}
\begin{minipage}[b]{0.48\linewidth}
\centering
\vspace{18pt}
\includegraphics[width=\textwidth]{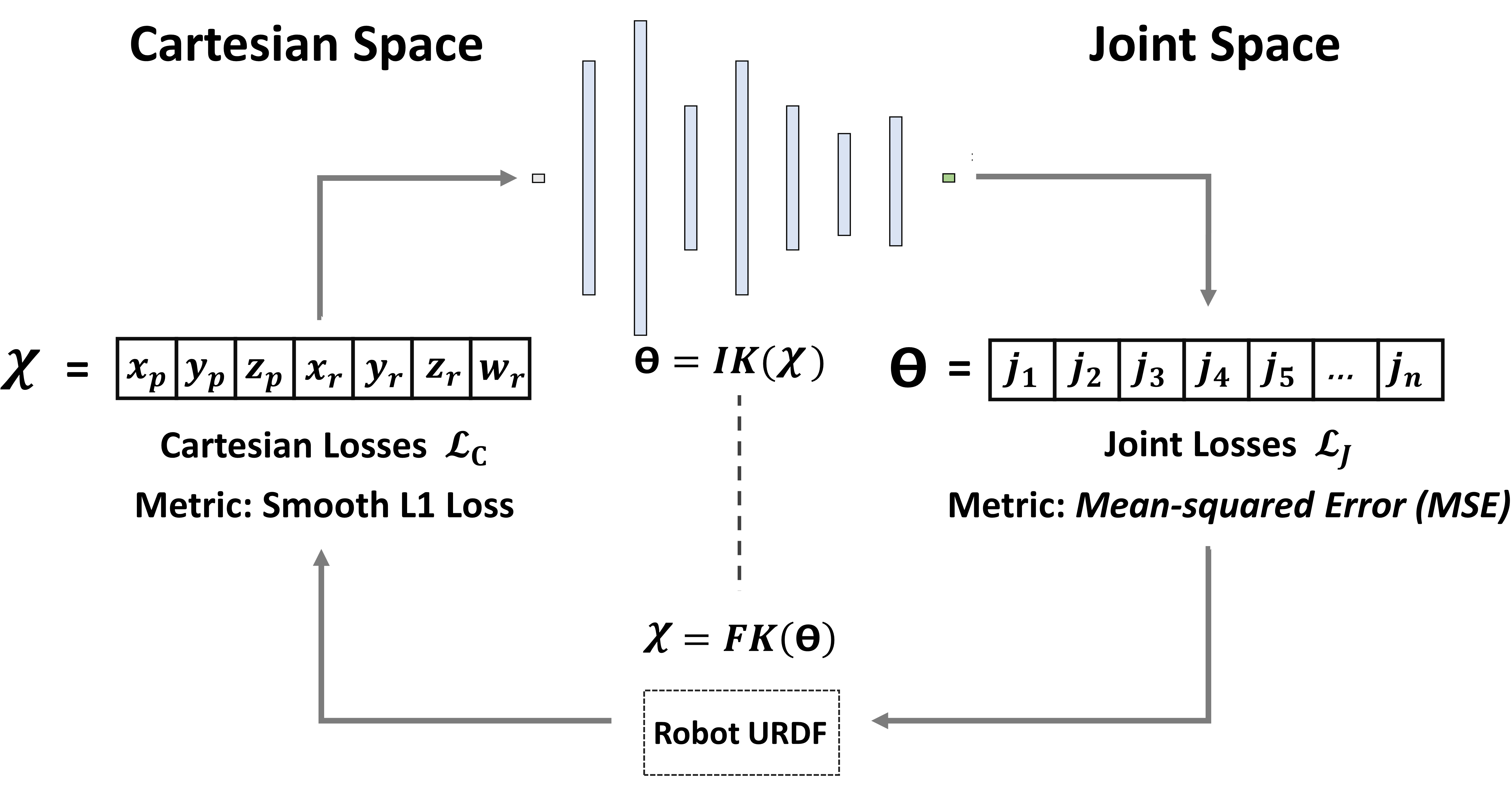}
\vspace{2pt}
\caption{Overview of the CycleIK training algorithm. Optional secondary kinematic constraints can either be set in the joint space or Cartesian space.}
\label{fig:cycleik_overview}
\end{minipage}
\label{fig:architecture_training}
\end{figure*}

\subsection{Architecture}

Our approach relies on a hyperparameter-optimized fully connected network in the form of an MLP (Fig. \ref{fig:cycleik_architecture}) that quickly learns to approximate the inverse kinematics transformation from Cartesian to joint space. The networks have between six to eight layers. The exact number of layers and neurons per layer are part of the optimized hyperparameters.   

The input to the networks is a batch of one-dimensional vectors with a length of seven that represents the 6D pose of the robot end effector. Each of the vectors in the input batch holds three fields for the Cartesian position and four fields for the rotation in the quaternion form. The output domain of the MLP is the joint space of the robot, which is represented as a one-dimensional vector with the length of the robot DoF and thus varies for different robot designs.

Both, the input and the output are normalized into the interval $[-1, 1]$. The output is limited to the interval by design as it has the $\tanh$ activation. The input and hidden layers are Gaussian-Error Linear Units (GELUs)\cite{Hendrycks2016GaussianGELUs} and are thus not limited. The input is normalized to the specified interval by the data loader so that our architecture does not functionally restrict the input values to the noted interval and likely produces unexpected behavior for inputs that lie out of the interval and consequently the training data distribution.
  
\subsection{Training}
The fundamental idea of the training algorithm is shown in Fig.~\ref{fig:cycleik_overview}. The proposed model learns to solve the IK problem by training on data batch $\mathcal{X}$ that purely consists of Cartesian poses. The training algorithm most importantly makes use of the inverse property between the $\mathrm{FK}(\Theta)$ and $\mathrm{IK}(\mathcal{X})$ function, as noted in Eq.\ref{eq:ik_fk_inverse}:

\begin{equation}
   \mathcal{X} = \mathrm{FK}(\Theta) \hspace{3pt} \land \hspace{3pt} \Theta = \mathrm{IK}(\mathcal{X})
   \label{eq:ik_fk_inverse}
\end{equation}

To avoid the labeling of poses with pre-calculated joint angles, which performs poorly for redundant manipulators like the NICOL robot\cite{Kerzel2023NICOL:Manipulation}, the joint angles inferred from the MLP $\Theta$ are transformed back to Cartesian space. The Cartesian distance is calculated between the input pose and the end effector pose derived from the IK model, as shown in Eq.~\ref{eq:cycle_infer}:

\begin{equation}
   \mathcal{X} \approx \hat{\mathcal{X}} = \mathrm{FK}(\mathrm{IK}(\mathcal{X}))
    \label{eq:cycle_infer}
\end{equation}

We utilize the geometric forward kinematics solver {\textit{PyTorch Kinematics}}\cite{Zhong2023PyTorchKinematics} that allows for batched FK queries as it calculates the end effector pose for a particular joint state by matrix multiplications and allows for computational acceleration by utilizing the GPU via PyTorch. The operations used by {\textit{PyTorch Kinematics} are differentiable and can therefore be used during the network training without blocking the propagation of the gradient. Every training iteration processes a batch of poses with a minimum size of $100$, while the exact batch size is a hyperparameter that is optimized per robot. The learning rate is linearly decreased to zero over the training time. 


\subsection{Metrics}
The updated loss functions proposed by this paper utilize the smooth L1 loss as the central metric to calculate the positional and rotational error.
The smooth L1 metric\cite{Girshick2015FastR-CNN} is a parameterized variant of the well-known Huber loss\cite{Huber1964RobustParameter}. Huber loss was used in various machine learning applications and has also shown good performance for gradient-based systems\cite{Friedman2001GreedyMachine.}.

The former positional loss function of CycleIK calculated the mean absolute error (MAE) over the batch of target positions $P \coloneqq \{p_0, ..., p_n\}$ and predicted result positions $\hat{P}  \coloneqq \{\hat{p}_0, ..., \hat{p}_n\}$. Given a batch of size $N$, the former loss function $\mathcal{L}_{pos}$ can be defined as in Eq.~\ref{eq:old_position_loss}:

\begin{equation}
   \mathcal{L}_{pos} = \frac{1}{N}\sum_{i=1}^{N}\frac{1}{3}\sum_{j=1}^{3} |\hat{p}_{i,j} - p_{i,j}|
   \label{eq:old_position_loss}
\end{equation}

The MAE can result in steep gradients when the loss is near to zero. The new loss metric $\mathcal{L}_{pos}^\prime$ (Eq.~\ref{eq:new_position_loss}) for the positional error calculates the smoothed L1 distance $l(a, b)$ between position $p_n$ and prediction $\hat{p}_n$ as in \cite{Girshick2015FastR-CNN}:

\begin{equation}
   \mathcal{L}_{pos}^\prime = \frac{1}{N}\sum_{i=1}^{N}\frac{1}{3}\sum_{j=1}^{3} l(\hat{p}_{i,j}, p_{i,j})
   \label{eq:new_position_loss}
\end{equation}

First, the average over the three axes of the position and following the batch mean is calculated. The smooth L1 loss, compared to the Huber loss, introduces an additional parameter $\beta$. According to \cite{Girshick2015FastR-CNN}, when $\beta = 1$, the smooth L1 loss is equal to the Huber loss. For the positional error, whose unit is meters, we choose $\beta = 0.001$, which means that errors below $1\,\mathrm{mm}$ will be smoothed. Our notion of the smooth L1 loss $l(a, b)$ is based on the definition in the PyTorch documentation\footnote[3]{https://pytorch.org/docs/stable/generated/torch.nn.SmoothL1Loss.html} as shown in Eq.~\ref{eq:smoothl1_loss}: 

\begin{equation}
 l(a, b) = \begin{cases}
\tfrac{1}{2}(a - b)^2/\beta, &\text{if $|a - b| < \beta$}\\
|a - b| - \tfrac{1}{2} \beta, &\text{otherwise}
\end{cases}
\label{eq:smoothl1_loss}
\end{equation}

Similar to the former positional loss metric, the prior rotational loss metric also calculated the MAE between the target quaternion $q_n$ and the quaternion $\hat{q}_n$ reached by the MLP. The old rotational loss function $\mathcal{L}_{rot}$ of CycleIK for a batch of size $N$ is stated in Eq.~\ref{eq:old_rotation_loss}:  

\begin{equation}
   \mathcal{L}_{rot} = \frac{1}{N}\sum_{i=1}^{N}\frac{1}{4}\sum_{j=1}^{4} |\hat{q}_{i,j} - q_{i,j}|
   \label{eq:old_rotation_loss}
\end{equation}

However, the loss metric from Eq.~\ref{eq:old_rotation_loss} is problematic from multiple perspectives. Two quaternions exist for every possible orientation in spherical SO(3) space due to the existence of the positive and negative imaginary part of the complex orientation representation. The metric from Eq.~\ref{eq:old_rotation_loss} assumes that the FK solver that is used for the generation of the data set and the FK solver used during training, deterministically return the same either positive or negative imaginary part for a given joint state. According to\cite{Huynh2009MetricsAnalysis}, the ambiguous quaternion of a quaternion $q \in$ SO(3) can be calculated by the negation of $q$, as noted in Eq.~\ref{eq:quaternion_ambiguity}:    

\begin{equation}
   q_i = -q_i \hspace{5pt}\text{for}\hspace{5pt} q_i, -q_i \in SO(3)
   \label{eq:quaternion_ambiguity}
\end{equation}

Given two very similar orientations $q_1, -q_2: q_1 \approx -q_2$ are contained in the training data set, the gradients will likely explode when both samples are contained in the same training batch and the metric from Eq.~\ref{eq:old_rotation_loss} is used.

We introduce a Smooth Minimum Quaternion Loss (SMQL) in Eq.~\ref{eq:smql_loss} based on the smooth L1 loss and the $\phi_2$ metric for rotations proposed in \cite{Huynh2009MetricsAnalysis} to handle the ambiguity property of quaternions:

\begin{equation}
   \mathcal{L}_{rot}^\prime = \frac{1}{N}\sum_{i=1}^{N}\min\{\sum_{j=1}^{4} l(\hat{q}_{i,j}, q_{i,j}), \sum_{j=1}^{4} l(\hat{q}_{i,j}, -q_{i,j})\}
   \label{eq:smql_loss}
\end{equation}

First, the element-wise smooth L1 loss $l(\hat{q_i}, q_i)$, with $\beta = 0.01$ corresponding to $\sim1^{\circ}$, is calculated over the batch of target quaternions $Q  \coloneqq \{q_0, ..., q_n\}$ and the predicted orientations $\hat{Q}  \coloneqq \{\hat{q}_0, ..., \hat{q}_n\}$. The same procedure is repeated for the negated target orientations $-Q$ so that $l(\hat{q_i}, -q_i)$ is calculated. The resulting errors of shape $(N \times 4)$ are summed up row-wise. Then the minimum between the distance to the positive $q_i$ and the negative target rotation $-q_i$ is chosen for every index $i$. Finally, the batch mean is calculated over the resulting batch of minimum sums of errors.

Our network is consequently trained by a multi-objective function that calculates the positional error $\mathcal{L}_{pos}^\prime$ according to Eq.~\ref{eq:new_position_loss} and the rotational error $\mathcal{L}_{rot}^\prime$ according to Eq.~\ref{eq:smql_loss} to consequently build a weighted sum with the optional set of secondary constraints $\mathcal{L}_{secondary}$ and their weights $\mathcal{W}_{secondary}$ as noted in Eq.~\ref{eq:multi_objective_function}:

\begin{equation}
\begin{split}
   \mathcal{L} = w_{pos} * \mathcal{L}_{pos}^\prime + w_{rot}& * \mathcal{L}_{rot}^\prime + \sum_{k=1}^{K} w_{k} * \mathcal{L}_{k} \\
   w_k \in \mathcal{W}_{secondary}, \hspace{3pt} & \hspace{3pt} \mathcal{L}_k \in \mathcal{L}_{secondary}
\end{split}   
\label{eq:multi_objective_function}
\end{equation}

Previously, we reported issues with the stability of the old method\cite{10.1007/978-3-031-44207-0_38}. Our system became stable by introducing the smooth L1 metric for the positional and SMQL for the rotational error. The reported issues were not encountered anymore in any of the preliminary experiments for this publication. The variance of the test error for multiple runs is small.

\subsection{Dataset}
\label{sec:datasets}
The datasets that are used consist purely of Cartesian end effector poses of the corresponding robots. The datasets are created from uniform random samples in the joint space of the robot. Samples that are in a collision state are filtered out. The Moveit\cite{Coleman2017ReducingStudy} FK and collision detection functionalities are used for the data set aggregation. The approach independently generates training, validation, and test sets for each robot. The training data sets contain one million samples each, while the validation sets contain $100,000$ and the test sets $200,000$ samples each.

\begin{table*}[t!]
\vspace{4pt}
        \begin{center}

        \caption{Hyperparameter configurations of the most successful evaluated trials for each robot.}
        \label{tab:optimized_hyper_params}
            \begin{tabular}{|c|cccccccc|}
                \hline
                Robot & Batch Size  & Lr$^a$ & No. layers & No. tanh & $w_{pos}$ & $w_{rot}$ &Number of neurons per layer& No. parameters \\
                \hline
                NICOL & $100$ & $1.8$  & $7$ & $1$ & $9$ & $2$& $[ 2780, 3480, 1710, 2880, 1750, 1090, 1470 ]$ &$29.146 * 10^6$\\
                NICO & $300$ & $3.7$ & $6$ & $1$ & $7$ & $1$& $[ 2270, 560, 1100, 1990, 2590, 870 ]$&$11.514 * 10^6$\\
                Valkyrie & $100$ & $4.4$ & $7$ & $1$ & $9$ & $1$ & $[ 2930, 1130, 1520, 570, 670, 770, 2250 ]$&$8.578 * 10^6$\\
                Panda & $100$ & $2.4$ & $7$ & $1$ & $16$ & $2$& $[ 1370, 880, 2980, 1000, 2710, 2290, 880 ]$&$17.767 * 10^6$\\
                Fetch & $100$ & $3.2$ & $7$ & $1$ & $19$ & $3$&  $[ 850, 620, 3210, 2680, 680, 3030, 2670 ]$&$23.134 * 10^6$\\
                \hline
                \multicolumn{9}{|l|}{\footnotesize{$^a$ Learning rate reported in units of $10^{-4}$}}\\
                \hline
            \end{tabular}\\
            \end{center}
            \label{tab:my_label}
    
\end{table*}

\subsection{Chat Agent and Object Detection}
The embodied chat agent introduced by Allgeuer et al.\cite{Allgeuer2024TBAAgent} is utilized to create a human-in-the-loop grasping scenario in which the user can interact with the NICO and NICOL robots via verbal instructions. The speech signal from the microphone is processed into natural language with OpenAI Whisper\cite{Radford2022RobustSupervision}. The open-vocabulary object detector ViLD\cite{Gu2021Open-vocabularyDistillation} is used to detect objects that are located on the table in front of the robots by processing the 2D image stream from the robot's eye cameras. Finally, OpenAI GPT-3.5\cite{OpenAI2024OpenAIgpt-3.5-turbo-0301} serves as the manager of the chat that processes the natural language from the recognized speech signal. The output of GPT is transformed to the frequency domain via text-to-speech. 
 
GPT is made aware of its role as a robot collaborator by an initial system prompt that introduces the main information about the assembly, the sensory inputs, and the physical abilities of the corresponding robot. The chat agent receives recurring information about the current objects on the table by appending the latest detection results to the user input in structured natural language. Similar to the embedding of the sensory input, GPT is allowed to embed action tags in the form of structured natural language in its output, which enables the system to execute actions in the physical world, e.g. face expressions, pointing gestures, and pushing primitives. We foster robotic platform independence by generalizing our robotic agent, that now can not only be embodied by NICOL but also by NICO. We contribute a grasping primitive to the agents' action space, that allows for precise manipulation.

\subsection{Motion Planning}
Grasping motions can quickly be generated by a small set of IK queries that are dynamically adapted towards a target position, while always maintaining the same orientation\cite{Kerzel2020Neuro-GeneticGrasping, Kerzel2023NICOL:Manipulation}, \cite{Gaede2024}. However, the approach lacks scalability to other robot workspaces and can lead to unnatural movements that, in the worst case, include dynamic singularities when the velocity and acceleration profiles of the physical motors are insufficiently tuned. Our approach utilizes B\'ezier curves for a motion planner. They are traditionally a popular motion planning method for Cartesian robots, e.g. sorting robots\cite{Riboli2023Collision-freeRepresentation}. The B\'ezier procedure, according to\cite{Zhao2023GlobalCurves}, interpolates a smooth trajectory in Cartesian space between a set of $n$ control points $C$, where $c_0$ is the start point and $c_n$ is the target point, as defined in Eq.~\ref{eq:control_points_bezier}: 

\begin{equation}
   C  \coloneqq \{c_0, ... \hspace{1pt}, c_n\}
   \label{eq:control_points_bezier}
\end{equation}

As a secondary property, the parameterization $C$\textbackslash$\{c_0, c_n\}$ of B\'ezier curves can also be learned with machine learning methods, as in\cite{Scheiderer2019BezierRobots}, but they are tuned manually in this study.

B\'ezier curves define a continuous trajectory when the steps between two points are infinitely small. Our method creates a set of $N$ discrete steps,  which are normalized to the interval $[0, 1]$, that are used to sample waypoints from the Cartesian trajectory, as in Eq.~\ref{eq:bezier_time_steps}:

\begin{equation}
   k = n * \frac{1}{N} \hspace{2pt}, \hspace{5pt} 0 \leq n \leq N, \hspace{2pt} n \in \mathbb{N}^+
   \label{eq:bezier_time_steps}
\end{equation}

Therefore, the variable $k$ can be seen as an abstract measure of time where $0$ marks the starting and $1$ the end point of the motion. B\'ezier curves interpolate a single point $p_{plan}^k$ on the discrete trajectory by weighting the influence of each control point in $C$ conditioned to $k$. We adopt the notion for B\'ezier curves from \cite{Zhao2023GlobalCurves} as noted in Eq.~\ref{eq:bezier_notion}: 

\begin{equation}
   \begin{split}
   p_{plan}^k = \sum_{i = 0}^{t} \binom{t}{i}& (1 - k)^{t-i} k^i c_i \hspace{2pt} \\
   k \in [0, 1] \subset \mathbb{R}& \hspace{2pt}, \hspace{5pt} t = \Bar{C}
\end{split}  
\label{eq:bezier_notion}
\end{equation}

Similarly, Slerp\cite{Shoemake1985AnimatingCurves} is used to generate minimum-torque spherical trajectories by interpolating cubic B\'ezier curves between start quaternion $q_{start}$ and target quaternion $q_{target}$ in the spherical SO(3) space. Slerp can also handle a set of control points $C$ in the interpolation. For every key rotation $c_n$ and the following key rotation $c_{n+1}$, a cubic B\'ezier trajectory is interpolated according to Eq.~\ref{eq:bezier_notion}, which results in a number of $n-1$ partial B\'ezier trajectories in SO(3) for $n$ key rotations.

A grasping primitive was implemented that grasps an object on the table in front of the robots and deposits it into a bin that is located on either the right- or left-hand side towards the outer workspace. A visualization of a planned trajectory is given in Fig.~\ref{fig:vis_trajectory}. The object position is transformed from the image plane to the 3D coordinate through the extrinsic camera model. A B\'ezier-based motion plan is generated towards the 3D target position, with a fixed top grasp orientation at the target. The Cartesian B\'ezier curve, a batch that holds $50$ poses, is transformed into a joint trajectory through the CycleIK MLP in a single step. We configured the control points of the motion planner to approach the target objects from above, which matches with the target orientation at the object. When the execution of the motion by the manipulator is finished, the tendon-driven fingers close to create a form-closed grasp. A second motion that lifts the object and deposits it into one of the bins is planned and executed before the robot moves back to its idle configuration. The runtime of the Cartesian planner lies around $70\,\mathrm{ms}$ for trajectories with $50$ points on average.

\section{Results}

\subsection{Hyperparameter Optimization}
The hyperparameters of our MLP models are optimized with the Optuna \cite{Akiba2019Optuna:Framework} framework. $200$ different hyperparameter configurations, that are sampled with a Tree-structured Partizan Estimator (TPE), are evaluated for each robot. We optimize the following hyperparameters: batch size, learning rate (Lr), number of layers, number of neurons per hidden layer, number of tanh (No. tanh) layers at the end of the network, position weight ($w_{pos}$), and orientation weight ($w_{rot}$). The evaluation of each configuration is conducted with a fixed weighting of the losses with a factor of $10$ for $w_{pos}$ and $1$ for $w_{rot}$. The parameter configurations of the most successful runs for each robot are shown in Table~\ref{tab:optimized_hyper_params}. 

\begin{figure}[t!]
      \centering
      \vspace{0pt}
      \includegraphics[width=0.87\linewidth]{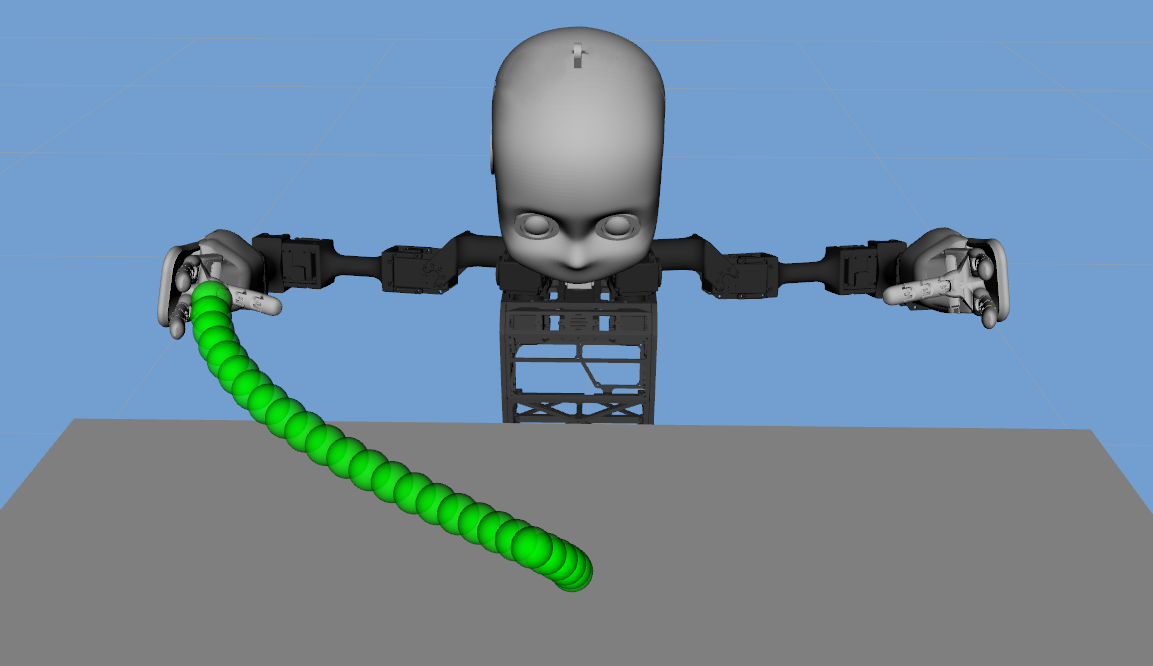}
      \caption{Visualization of the B\'ezier end effector trajectory for the right arm of the NICO robot.}
      \label{fig:vis_trajectory}
\end{figure}

\begin{table*}[t]
\vspace{4pt}
\caption{The precision of CycleIK compared to the numerical BioIK and Trac-IK methods. The positional (pos.) and rotational (rot.) error in $\mathrm{mm}$ and $^\circ$ and their corresponding standard deviations ($\sigma$), as well as the success rate (success) and the average runtime (time) in $\mathrm{ms}$ are reported.}
\label{tab:moveit_results}
\begin{center}
\setlength{\tabcolsep}{4.05pt}
\begin{tabular}{|c|c|c|c|c|c|c|c|c|c|c|c|c|c|c|c|c|c|c|}
\hline
Method & \multicolumn{6}{c|}{CycleIK} & \multicolumn{6}{c|}{BioIK} & \multicolumn{6}{c|}{Trac-IK}\\
\cline{1-1}
Robot & \multicolumn{1}{c}{pos.$^a$} & \multicolumn{1}{c}{$\sigma_{pos}$} &\multicolumn{1}{c}{rot.$^b$} & \multicolumn{1}{c}{$\sigma_{rot}$} & \multicolumn{1}{c}{success} & time$^c$& \multicolumn{1}{c}{pos.$^a$} & \multicolumn{1}{c}{$\sigma_{pos}$} &\multicolumn{1}{c}{rot.$^b$} & \multicolumn{1}{c}{$\sigma_{rot}$} & \multicolumn{1}{c}{success} & time$^c$& \multicolumn{1}{c}{pos.$^a$} & \multicolumn{1}{c}{$\sigma_{pos}$} &\multicolumn{1}{c}{rot.$^b$} & \multicolumn{1}{c}{$\sigma_{rot}$} & \multicolumn{1}{c}{success} & time$^c$\\
\hline
NICOL & 0.94 & \textbf{2.17}& 0.24 & \textbf{0.93}& 99.23\%& \textbf{0.39}& \textbf{0.54} & 15.6& \textbf{0.09} & 2.75& \textbf{99.87\%}& 1.00& 6.62 & 54.7& 1.10 & 9.33& 98.49\%& 1.00\\
NICO & \textbf{2.44} & \textbf{5.28}& \textbf{0.70} & \textbf{1.74}& \textbf{97.31\%}& \textbf{0.36}& 24.1 & 85.2& 5.72 & 21.3 & 92.10\%& 1.00& 20.7 & 79.6& 4.92 & 19.8 & 93.21\% & 1.00\\
Valkyrie & \textbf{2.01} & \textbf{6.98}& 0.63 & \textbf{2.29}& 97.33\%& \textbf{0.39}& 2.10 & 29.7 & \textbf{0.43} & 6.0& \textbf{99.41\%}& 1.00& 66.9 & 150& 13.8 & 30.5& 80.52\%& 1.00\\
Panda & \textbf{5.70} & \textbf{11.3}& \textbf{1.95} & \textbf{4.24}& 89.18\%& \textbf{0.39}& 12.2 & 71.2& 2.53 & 14.5& \textbf{96.76\%}& 1.00& 44.0 & 125& 9.56 & 26.7& 87.46\%& 1.00\\
Fetch & 3.11 & \textbf{9.14}& 0.79 & 2.30& 94.86\%& \textbf{0.38}& \textbf{0.35} & 12.8& \textbf{0.05} & \textbf{1.92}& \textbf{99.91\%}& 1.00& 13.4 & 81.4& 2.09 & 12.9& 97.19\%& 1.00\\
\hline
\multicolumn{19}{|l|}{\footnotesize{$^a$ Mean positional error and corresponding standard deviation reported in $\mathrm{mm}$}}\\
\multicolumn{19}{|l|}{\footnotesize{$^b$ Mean rotational error and corresponding standard deviation reported in degree}}\\
\multicolumn{19}{|l|}{\footnotesize{$^c$ Algorithm runtime per single sample reported in $\mathrm{ms}$}}\\
\hline
\end{tabular}
\end{center}
\end{table*}

\begin{table}[t]
\caption{Overview of the precision of SOTA neuro-inspired IK solvers CycleIK, IKFlow, and EEM. The positional error (pos.) and rotational error (rot.) are reported in $\mathrm{mm}$ and $^\circ$ and the runtime (time) on a batch of $100$ poses in $\mathrm{ms}$.}
\label{tab:SOTA_results}
\begin{center}
\setlength{\tabcolsep}{5.6pt}
\begin{tabular}{|c|c|c|c|c|c|c|c|c|}
\hline
Method & \multicolumn{3}{c|}{CycleIK} & \multicolumn{3}{c|}{IKFlow\cite{Ames2022IKFlow:Solutions}} & \multicolumn{2}{c|}{EEM$^a$\cite{Limoyo2023EuclideanKinematics}}\\
\cline{1-1}
Robot & \multicolumn{1}{c}{pos.} &\multicolumn{1}{c}{rot.}&time & \multicolumn{1}{c}{pos.} & \multicolumn{1}{c}{rot.}&time& \multicolumn{1}{c}{pos.} & rot. \\
\hline
Panda & \textbf{5.70}& 1.95& \textbf{0.41}&  7.72& 2.81& 8.55&  12.30& \textbf{1.00}\\
Valkyrie & \textbf{2.01}& \textbf{0.63}& \textbf{0.43}& 3.49& 0.74& 6.28& --& --\\
\hline
\multicolumn{9}{|l|}{\footnotesize{$^a$ Runtime not reported in publication}}\\
\hline
\end{tabular}
\end{center}
\end{table}

\subsection{Precision}

The neuro-inspired inverse kinematics method CycleIK is evaluated on five different robot platforms in simulation: the NICOL\cite{Kerzel2023NICOL:Manipulation}, NICO\cite{Kerzel2017NICO-Neuro-inspiredInteraction}, NASA Valkyrie\cite{Radford2015Valkyrie:Robot}, Franka Panda\cite{Haddadin2022TheEducation}, and Fetch\cite{Wise2016FetchApplications} robot. No secondary constraints were set for the training. We evaluate the genetic algorithm BioIK\cite{Starke2017AMotion} and the Jacobian-based Trac-IK\cite{Beeson2015TRAC-IK:Kinematics} on the same test data sets for a baseline. Both numerical IK algorithms are available as a plug-in in Moveit\cite{Coleman2017ReducingStudy}. The results of our experiments can be seen in Table~\ref{tab:moveit_results}. Our test data sets contain $200,000$ poses for each robot that were randomly sampled in the joint space of the robots with Moveit (See Sec.~\ref{sec:datasets}). 

We report the mean positional and rotational error as well as the corresponding standard deviation of the error in millimeters and degrees. MoveIt is allowed to return in-collision solutions, as collision-freeness is also not guaranteed by CycleIK. The algorithms are allowed a runtime of $1\,\mathrm{ms}$, which we define to be the minimum required for Moveit. The error cannot be evaluated when it exceeds an internal Moveit threshold, as no solution is returned by the algorithms. We calculate the error between the idle configuration of the corresponding robot and the target pose in these cases. We utilize the success definition of Kerzel et al.\cite{Kerzel2020Neuro-GeneticGrasping, Kerzel2023NICOL:Manipulation} which allows for $10\,\mathrm{mm}$ positional and $20\,^\circ$ of rotational error.

Overall, it can be seen that CycleIK always outperforms BioIK and Trac-IK in terms of the algorithm runtime. BioIK has most often the highest success rate, except for the NICO robot test set. BioIK shows the lowest rotational errors for three of the five environments (NICOL, Valkyrie, Fetch), while CycleIK shows the lowest positional error for three of the five robots (NICO, Valkyrie, Panda). Trac-IK is the least precise and successful method in our experiments and is only capable of outperforming BioIK on the NICO robot and CycleIK regarding the success rate on the Fetch robot. CycleIK always shows the lowest spread of the error distribution, in terms of the standard deviation of the error. The error distribution of Trac-IK always shows the highest variance except for the NICO robot where BioIK has the highest. BioIK only has the lowest standard deviation of the rotational error on the Fetch test set. The superiority of CycleIK in this metric can be explained by the fact that neural networks generalize well over the domain they were trained in so that a lot of unsuccessful solutions will lie only slightly above the limits of our error definition and thus result in a lower variance. For Moveit solvers like BioIK and Trac-IK, the variance is intuitively higher, as no solutions that lie out of the Moveit success definition can be evaluated so that the distance to the idle pose must be calculated which automatically increases the variance of the error.

Furthermore, Table~\ref{tab:SOTA_results} provides an overview of state-of-the-art neuro-inspired IK methods, including IKFlow\cite{Ames2022IKFlow:Solutions} and the Euclidean Equivariant Models (EEMs) proposed by Limoyo et al.\cite{Limoyo2023EuclideanKinematics}. The experimental results for IKFlow and EEM are taken from the corresponding publication. The mean positional and mean rotational error are the only metrics reported in both other papers. Limoyo et al. do not report the inference time. IKFlow reports the average runtime over batches that contain $100$ poses each. We adopted our experiment to this setup and always inferred the IK solution for a batch that holds $100$ poses, which results in a slightly higher runtime compared to the results that are reported in Table~\ref{tab:moveit_results}, but the same positional and rotational precision as the MLP models deterministically return the same solution for the identical pose, independent of the batch size.

CycleIK shows the overall highest precision and is only outperformed by the EEM approach regarding the rotational error on the Franka Panda robot test set. It has to be noted that IKFlow and EEMs are generative approaches that are capable of sampling the pose null space, while the CycleIK MLP deterministically returns one solution for every pose. In addition, EEMs are a multi-robot IK method that is capable of solving the IK problem for five different robots with a single model. Consequently, it has to be acknowledged that the complexity of the task is thus strongly relieved for the CycleIK MLP when compared to the IKFlow and EEM approaches. The increase of the batch size from $1$ to $100$ during inference only had a small effect on the average inference time per batch, the runtime for the Franka Panda robot increased by $0.02\,\mathrm{ms}$ to $0.41\,\mathrm{ms}$, and for the Valkyrie robot by $0.05\,\mathrm{ms}$ to $0.43\,\mathrm{ms}$. Overall, batch sizes up to $1000$ on average did not increase the runtime above $1\,\mathrm{ms}$ in our preliminary tests. While a direct comparison is complex, all three models present a new class of neural IK methods that reach a millimeter range of error compared to centimeters as in many previous approaches in this field, e.g. \cite{Kim2021LearningManipulator} or \cite{Lembono2020LearningNetwork}.

\subsection{Grasping}
The grasping primitive previously described in this work is deployed to the physical NICO and NICOL hardware with the chat agent from Allgeuer et al.\cite{Allgeuer2024TBAAgent}, which is expanded to be embodied by both humanoids. The Cartesian action space for the grasping primitive is an area of $40\,\mathrm{cm}$ width and $10\,\mathrm{cm}$ depth on the table in front of the NICO robot. The action space on the table in front of the NICOL robot is an $80\,\mathrm{cm}$ by $30\,\mathrm{cm}$ area, which is more than twice as large as for NICO. Each robot was tasked to perform $100$ grasps with small sets of objects that were individually selected for each of the robots. 

Five household objects from the YCB dataset\cite{CalliYCB} were used for the NICOL robot: a tomato soup can, a white baseball, a yellow banana, a green pear, and an orange (See Fig.~\ref{fig:nicol}). For the NICO robot, three differently shaped plush toys were used that all had a size of around $4\,\mathrm{cm}$: a bunch of purple grapes, a red tomato, and a halved lemon. We always arranged all objects of the set on a horizontal line that is spread over the width of the table. We moved the line in a step size of $2\,\mathrm{cm}$ for NICO and $5\,\mathrm{cm}$ for NICOL over the depth coordinate of the action space so that the whole action area was covered. For every row, the experimenter ensured that the object detector ViLD, which tends to oscillate over different object classes, had stabilized before verbally instructing the agent to execute the grasps. A wrong prediction of the object class that resulted in a successful grasp trial was not considered a failure. The agent was given the same verbal instruction for every row: \textit{'\{NICO $\vert$ NICOL\}, please grasp all the objects on the table, one after another. Choose the order of the objects yourself.'} 

\begin{figure}[t!]
      \vspace{5pt}
      \centering
      \includegraphics[width=1.0\linewidth]{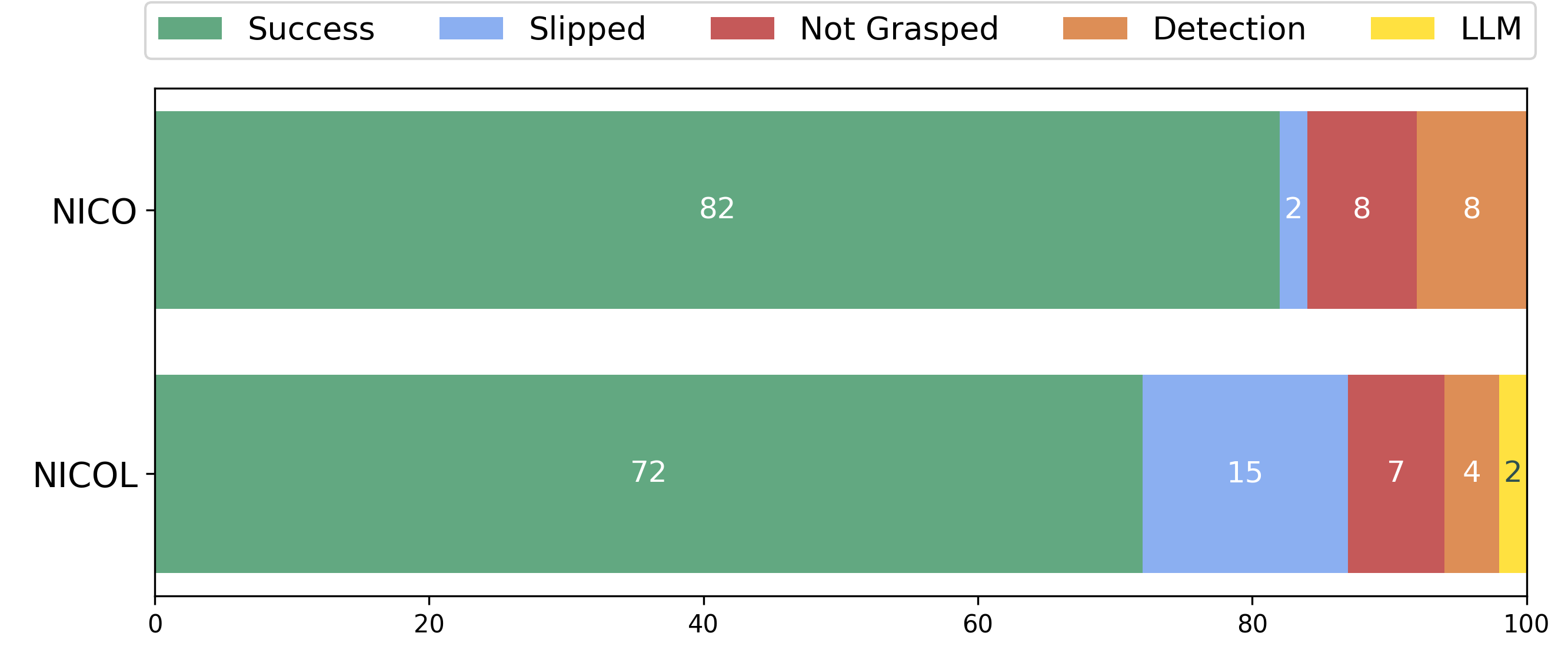}
      \caption{The error distribution of the B\'ezier-based motion planner in a human-in-the-loop grasping scenario.}
      \label{fig:grasping_success}
\end{figure}


The results of our experiment can be seen in Fig.~\ref{fig:grasping_success}. We defined five different result classes into which every grasp approach is classified during the experiment: \textit{Success}, \textit{Slipped}, \textit{Not Grasped}, \textit{Detection}, and \textit{LLM}. A grasp approach is considered \textit{Success} when the object is lifted from the table and correctly disposed of into one of the bins. An approach is considered \textit{Slipped} when it is correctly lifted from the table vertically but slips out of the hand during the transfer to the bin. Approaches that fail to grasp the object correctly and that are not able to lift it from the table are classified as \textit{Not Grasped}. The \textit{Detection} class contains approaches that cannot successfully be executed because the object detector predicts a wrong or no class at all, or a completely wrong location. The \textit{LLM} class contains approaches in which the language model is aware of all the objects on the table but does not execute the grasp action on all of them.

Our method achieves $72\%$ success on the NICOL and $82\%$ on the NICO robot. The objects slipped out of NICOL's robot hands in 15 trials, while the same only happened twice on the NICO robot and therefore poses the most significant difference between both robots. Some possible reasons for the performance difference are that the household objects that NICOL interacts with are heavier, not deformable, and have less friction. Therefore, NICOL´s task is more difficult. The number of completely failed grasps shows a very similar error ratio on both robots with eight missed grasps on the NICO robot and seven on the NICOL robot. The object detection worked twice as reliably on the NICOL robot with four wrong detections as on the NICO robot with eight wrong detections. On  NICOL, the LLM did not consider grasping an object even though it was aware that the object was on the table in two cases.

\section{Conclusions}
This work successfully introduced a novel B\'ezier-based non-collision-free motion planner that is based on the neuro-inspired inverse kinematics solver CycleIK. New loss functions for the positional and rotational error calculation were introduced to CycleIK that were derived from the smooth L1 loss and enable high precision within a short training time. We have shown that the new CycleIK method can compete with and partially outperforms well-known numerical approaches such as BioIK and Trac-IK, but also state-of-the-art neuro-inspired approaches like IKFlow in terms of precision. We have shown that CycleIK is one of the fastest batched IK methods available and offers quick planning of robot motions in Cartesian space. The best iterative numerical approach performs better than CycleIK but it requires a longer runtime. However, CycleIK provides an ideal choice, when the runtime is a major critical factor and performs very good for the NICO and NICOL robots.  The motion planner was deployed to the physical humanoid robots NICO and NICOL in an embodied agent setup. Our method reached a success rate of $72\%$ on the NICOL robot and $82\%$ on the NICO robot in a human-in-the-loop grasping scenario.

\section*{Acknowledgement}
The authors thank Erik Strahl and Matthias Kerzel for
discussions on inverse kinematics for CycleIK;  Ozan Özdemir for camera development of the NICO robot;
Hassan Ali for his contributions to the object detection on the NICOL robot.

\addtolength{\textheight}{-0.2cm}

\bibliographystyle{IEEEtran}
\bibliography{cycleik}

\begin{thebibliography}{10}
\providecommand{\url}[1]{#1}
\csname url@samestyle\endcsname
\providecommand{\newblock}{\relax}
\providecommand{\bibinfo}[2]{#2}
\providecommand{\BIBentrySTDinterwordspacing}{\spaceskip=0pt\relax}
\providecommand{\BIBentryALTinterwordstretchfactor}{4}
\providecommand{\BIBentryALTinterwordspacing}{\spaceskip=\fontdimen2\font plus
\BIBentryALTinterwordstretchfactor\fontdimen3\font minus \fontdimen4\font\relax}
\providecommand{\BIBforeignlanguage}[2]{{%
\expandafter\ifx\csname l@#1\endcsname\relax
\typeout{** WARNING: IEEEtran.bst: No hyphenation pattern has been}%
\typeout{** loaded for the language `#1'. Using the pattern for}%
\typeout{** the default language instead.}%
\else
\language=\csname l@#1\endcsname
\fi
#2}}
\providecommand{\BIBdecl}{\relax}
\BIBdecl

\bibitem{MorganCppFlow:Planning}
J.~Morgan, D.~Millard, and G.~S. Sukhatme, ``Cppflow: Generative inverse kinematics for efficient and robust cartesian path planning,'' in \emph{2024 IEEE International Conference on Robotics and Automation (ICRA)}, 2024, pp. 12\,279--12\,785.

\bibitem{Park2022NODEIK:Planning}
S.~Park, M.~Schwartz, and J.~Park, ``{NODEIK: Solving Inverse Kinematics with Neural Ordinary Differential Equations for Path Planning},'' in \emph{2022 22nd International Conference on Control, Automation and Systems (ICCAS)}, 2022, pp. 944--949.

\bibitem{Qureshi2021MotionPlanners}
A.~H. Qureshi, Y.~Miao, A.~Simeonov, and M.~C. Yip, ``{Motion Planning Networks: Bridging the Gap Between Learning-Based and Classical Motion Planners},'' \emph{IEEE Transactions on Robotics}, vol.~37, no.~1, pp. 48--66, 2021.

\bibitem{10.1007/978-3-031-44207-0_38}
J.-G. Habekost, E.~Strahl, P.~Allgeuer, M.~Kerzel, and S.~Wermter, ``{CycleIK: Neuro-inspired Inverse Kinematics},'' in \emph{Artificial Neural Networks and Machine Learning – ICANN 2023, Proceedings, Part I}.\hskip 1em plus 0.5em minus 0.4em\relax Berlin, Heidelberg: Springer-Verlag, 2023, p. 457–470.

\bibitem{Girshick2015FastR-CNN}
R.~Girshick, ``{Fast R-CNN},'' in \emph{2015 IEEE International Conference on Computer Vision (ICCV)}, 2015, pp. 1440--1448.

\bibitem{Kerzel2023NICOL:Manipulation}
M.~Kerzel, P.~Allgeuer, E.~Strahl, N.~Frick, J.-G. Habekost, M.~Eppe, and S.~Wermter, ``{NICOL: A Neuro-Inspired Collaborative Semi-Humanoid Robot That Bridges Social Interaction and Reliable Manipulation},'' \emph{IEEE Access}, vol.~11, pp. 123\,531--123\,542, 2023.

\bibitem{Allgeuer2024TBAAgent}
P.~Allgeuer, H.~Ali, and S.~Wermter, ``{When Robots Get Chatty: Grounding Multimodal Human-Robot Conversation and Collaboration},'' University of Hamburg, Tech. Rep., 2024.

\bibitem{Kerzel2017NICO-Neuro-inspiredInteraction}
M.~Kerzel, E.~Strahl, S.~Magg, N.~Navarro-Guerrero, S.~Heinrich, and S.~Wermter, ``{NICO — Neuro-inspired companion: A developmental humanoid robot platform for multimodal interaction},'' in \emph{2017 26th IEEE International Symposium on Robot and Human Interactive Communication (RO-MAN)}, 2017, pp. 113--120.

\bibitem{OpenAI2024OpenAIgpt-3.5-turbo-0301}
\BIBentryALTinterwordspacing
{OpenAI}, ``{OpenAI GPT3.5 API [gpt-3.5-turbo-0301]},'' 2024. [Online]. Available: \url{https://platform.openai.com/docs/models/gpt-3-5-turbo}
\BIBentrySTDinterwordspacing

\bibitem{Gu2021Open-vocabularyDistillation}
X.~Gu, T.-Y. Lin, W.~Kuo, and Y.~Cui, ``{Open-vocabulary Object Detection via Vision and Language Knowledge Distillation},'' in \emph{International Conference on Learning Representations}, 2022.

\bibitem{Radford2022RobustSupervision}
A.~Radford, J.~W. Kim, T.~Xu, G.~Brockman, C.~McLeavey, and I.~Sutskever, ``{Robust speech recognition via large-scale weak supervision},'' in \emph{Proceedings of the 40th International Conference on Machine Learning}.\hskip 1em plus 0.5em minus 0.4em\relax JMLR.org, 2023.

\bibitem{Coleman2017ReducingStudy}
D.~T. Coleman, I.~A. Sucan, S.~Chitta, and N.~Correll, ``{Reducing the Barrier to Entry of Complex Robotic Software: a MoveIt! Case Study},'' \emph{Journal of Software Engineering for Robotics}, vol.~5, no.~1, pp. 3--16, 2017.

\bibitem{Starke2017AMotion}
S.~Starke, N.~Hendrich, and J.~Zhang, ``{A memetic evolutionary algorithm for real-time articulated kinematic motion},'' in \emph{2017 IEEE Congress on Evolutionary Computation (CEC)}, 2017, pp. 2473--2479.

\bibitem{Beeson2015TRAC-IK:Kinematics}
P.~Beeson and B.~Ames, ``{TRAC-IK: An open-source library for improved solving of generic inverse kinematics},'' in \emph{2015 IEEE-RAS 15th International Conference on Humanoid Robots (Humanoids)}, 2015, pp. 928--935.

\bibitem{Haddadin2022TheEducation}
S.~Haddadin~et al., ``{The Franka Emika Robot: A Reference Platform for Robotics Research and Education},'' \emph{IEEE Robotics and Automation Magazine}, vol.~29, no.~2, pp. 46--64, 6 2022.

\bibitem{Radford2015Valkyrie:Robot}
N.~A. Radford~et al., ``{Valkyrie: NASA's First Bipedal Humanoid Robot},'' \emph{Journal of Field Robotics}, vol.~32, no.~3, pp. 397--419, 5 2015.

\bibitem{Wise2016FetchApplications}
M.~Wise, M.~Ferguson, D.~King, E.~Diehr, and D.~Dymesich, ``{Fetch {\&} Freight : Standard Platforms for Service Robot Applications},'' Fetch Robotics, Tech. Rep., 2016.

\bibitem{Limoyo2023EuclideanKinematics}
O.~Limoyo, F.~Maric, M.~Giamou, P.~Alexson, I.~Petrovic, and J.~Kelly, ``{Euclidean Equivariant Models for Generative Graphical Inverse Kinematics},'' in \emph{RSS 2023 Workshop on Symmetries in Robot Learning}, 2023.

\bibitem{Ames2022IKFlow:Solutions}
B.~Ames, J.~Morgan, and G.~Konidaris, ``{IKFlow: Generating Diverse Inverse Kinematics Solutions},'' \emph{IEEE Robotics and Automation Letters}, vol.~7, no.~3, pp. 7177--7184, 2022.

\bibitem{Ardizzone2018AnalyzingNetworks}
L.~Ardizzone, J.~Kruse, C.~Rother, and U.~Köthe, ``{Analyzing Inverse Problems with Invertible Neural Networks},'' in \emph{International Conference on Learning Representations}, 2019.

\bibitem{Kim2021LearningManipulator}
S.~Kim and J.~Perez, ``{Learning Reachable Manifold and Inverse Mapping for a Redundant Robot manipulator},'' in \emph{2021 IEEE International Conference on Robotics and Automation (ICRA)}, 2021, pp. 4731--4737.

\bibitem{Lembono2020LearningNetwork}
T.~S. Lembono, E.~Pignat, J.~Jankowski, and S.~Calinon, ``{Learning Constrained Distributions of Robot Configurations With Generative Adversarial Network},'' \emph{IEEE Robotics and Automation Letters}, vol.~6, no.~2, pp. 4233--4240, 2021.

\bibitem{Bensadoun2022NeuralKinematic}
R.~Bensadoun, S.~Gur, N.~Blau, and L.~Wolf, ``{Neural Inverse Kinematic},'' in \emph{Proceedings of the 39th International Conference on Machine Learning}, ser. Proceedings of Machine Learning Research, vol. 162.\hskip 1em plus 0.5em minus 0.4em\relax PMLR, 17--23 Jul 2022, pp. 1787--1797.

\bibitem{Wagaa2023AnalyticalArm}
N.~Wagaa, H.~Kallel, and N.~Mellouli, ``{Analytical and deep learning approaches for solving the inverse kinematic problem of a high degrees of freedom robotic arm},'' \emph{Engineering Applications of Artificial Intelligence}, vol. 123, p. 106301, 2023.

\bibitem{Oreshkin2021ProtoRes:Kinematics}
B.~N. Oreshkin, F.~Bocquelet, F.~G. Harvey, B.~Raitt, and D.~Laflamme, ``{ProtoRes: Proto-Residual Network for Pose Authoring via Learned Inverse Kinematics},'' in \emph{International Conference on Learning Representations}, 2022.

\bibitem{Voleti2022SMPL-IK:Workflows}
V.~Voleti, B.~Oreshkin, F.~Bocquelet, F.~Harvey, L.-S. M\'{e}nard, and C.~Pal, ``{SMPL-IK: Learned Morphology-Aware Inverse Kinematics for AI Driven Artistic Workflows},'' in \emph{SIGGRAPH Asia 2022 Technical Communications}, ser. SA '22.\hskip 1em plus 0.5em minus 0.4em\relax New York, NY, USA: Association for Computing Machinery, 2022.

\bibitem{Li2021HybrIK:Estimation}
J.~Li, C.~Xu, Z.~Chen, S.~Bian, L.~Yang, and C.~Lu, ``{HybrIK: A Hybrid Analytical-Neural Inverse Kinematics Solution for 3D Human Pose and Shape Estimation},'' in \emph{2021 IEEE/CVF Conference on Computer Vision and Pattern Recognition (CVPR)}, 2021, pp. 3382--3392.

\bibitem{Li2023NIKI:Estimation}
J.~Li, S.~Bian, Q.~Liu, J.~Tang, F.~Wang, and C.~Lu, ``{NIKI: Neural Inverse Kinematics with Invertible Neural Networks for 3D Human Pose and Shape Estimation},'' in \emph{2023 IEEE/CVF Conference on Computer Vision and Pattern Recognition (CVPR)}, 2023, pp. 12\,933--12\,942.

\bibitem{Hendrycks2016GaussianGELUs}
D.~Hendrycks and K.~Gimpel, ``{Gaussian Error Linear Units (GELUs)},'' \emph{arXiv e-prints}, vol. arXiv:1606.08415, 6 2016.

\bibitem{Zhong2023PyTorchKinematics}
\BIBentryALTinterwordspacing
S.~Zhong, T.~Power, and A.~Gupta, ``{PyTorch Kinematics},'' 3 2023. [Online]. Available: \url{https://github.com/UM-ARM-Lab/pytorch{\_}kinematics}
\BIBentrySTDinterwordspacing

\bibitem{Huber1964RobustParameter}
P.~J. Huber, ``{Robust Estimation of a Location Parameter},'' \emph{The Annals of Mathematical Statistics}, vol.~35, no.~1, pp. 73 -- 101, 1964.

\bibitem{Friedman2001GreedyMachine.}
J.~H. Friedman, ``{Greedy function approximation: A gradient boosting machine.}'' \emph{The Annals of Statistics}, vol.~29, no.~5, pp. 1189 -- 1232, 2001.

\bibitem{Huynh2009MetricsAnalysis}
D.~Huynh, ``\BIBforeignlanguage{English}{{Metrics for 3D Rotations: Comparison and Analysis}},'' \emph{\BIBforeignlanguage{English}{Journal of Mathematical Imaging and Vision}}, vol.~35, no.~2, pp. 155--164, 2009.

\bibitem{Kerzel2020Neuro-GeneticGrasping}
M.~Kerzel, J.~Spisak, E.~Strahl, and S.~Wermter, ``{Neuro-Genetic Visuomotor Architecture for Robotic Grasping},'' in \emph{Artificial Neural Networks and Machine Learning – ICANN 2020, Proceedings, Part II}.\hskip 1em plus 0.5em minus 0.4em\relax Berlin, Heidelberg: Springer-Verlag, 2020, p. 533–545.

\bibitem{Gaede2024}
C.~Gaede, J.-G. Habekost, and S.~Wermter, ``Domain adaption as auxiliary task for sim-to-real transfer in vision-based neuro-robotic control,'' in \emph{Proceedings of the International Joint Conference on Neural Networks, Yokohama, Japan.}, 07 2024.

\bibitem{Riboli2023Collision-freeRepresentation}
M.~Riboli, M.~Jaccard, M.~Silvestri, A.~Aimi, and C.~Malara, ``{Collision-free and smooth motion planning of dual-arm Cartesian robot based on B-spline representation},'' \emph{Robotics and Autonomous Systems}, vol. 170, p. 104534, 2023.

\bibitem{Zhao2023GlobalCurves}
L.~Jia, S.~Zeng, L.~Feng, B.~Lv, Z.~Yu, and Y.~Huang, ``{Global Time-Varying Path Planning Method Based on Tunable Bezier Curves},'' \emph{Applied Sciences}, vol.~13, no.~24, 2023.

\bibitem{Scheiderer2019BezierRobots}
C.~Scheiderer, T.~Thun, and T.~Meisen, ``{Bézier Curve Based Continuous and Smooth Motion Planning for Self-Learning Industrial Robots},'' \emph{Procedia Manufacturing}, vol.~38, pp. 423--430, 2019, 29th International Conference on Flexible Automation and Intelligent Manufacturing (FAIM 2019).

\bibitem{Shoemake1985AnimatingCurves}
K.~Shoemake, ``{Animating rotation with quaternion curves},'' in \emph{Proceedings of the 12th Annual Conference on Computer Graphics and Interactive Techniques}, ser. SIGGRAPH '85, 1985, p. 245–254.

\bibitem{Akiba2019Optuna:Framework}
T.~Akiba, S.~Sano, T.~Yanase, T.~Ohta, and M.~Koyama, ``{Optuna: A Next-generation Hyperparameter Optimization Framework},'' in \emph{Proceedings of the 25th ACM SIGKDD International Conference on Knowledge Discovery \& Data Mining}, 2019, p. 2623–2631.

\bibitem{CalliYCB}
B.~Calli, A.~Singh, J.~Bruce, A.~Walsman, K.~Konolige, S.~Srinivasa, P.~Abbeel, and A.~M. Dollar, ``{Yale-CMU-Berkeley dataset for robotic manipulation research},'' \emph{The International Journal of Robotics Research}, vol.~36, no.~3, pp. 261--268, 2017.

\end{thebibliography}

\end{document}